\documentclass[9pt,journal,compsoc]{IEEEtran}

\ifCLASSOPTIONcompsoc
  \usepackage[nocompress]{cite}
\else
  \usepackage{cite}
\fi
\ifCLASSINFOpdf
\else
\fi

\usepackage{wrapfig,psfig,epsf}
\usepackage{todonotes}
\usepackage{xspace}
\usepackage{url}
\usepackage[inline,shortlabels]{enumitem}
\usepackage{graphicx}
\usepackage{booktabs}
\usepackage{multirow}  
\usepackage[T1]{fontenc}
\usepackage{amsmath}
\usepackage{subfig}
\usepackage{color}
\usepackage{xcolor}
\usepackage{colortbl}
\usepackage{amssymb}

\usepackage[nomessages]{fp}

\newcommand{\nirdi}{\texttt{Nirdizati}\xspace}
\newcommand{\ppm}{Predictive Process Monitoring\xspace}
\newcommand{\declare}{\textsc{declare}\xspace}
\newcommand{\fei}{\textit{Frequent Explanation Itemset}\xspace}
\newcommand{\feis}{\textit{Frequent Explanation Itemsets}\xspace}
\newcommand{\FEIs}{\ensuremath{\mathsf{FEIs}}\xspace}
\newcommand{\FEI}{\ensuremath{\mathsf{FEI}}\xspace}

\newcommand{\mscore}{$M$-score\xspace}

\newcommand{\simple}{\textit{simple-index}\xspace}
\newcommand{\complex}{\textit{complex-index}\xspace}

\newcommand{\act}[1]{\ensuremath{\mathsf{#1}}}
\newcommand{\att}[1]{\ensuremath{\mathsf{#1}}}

\newcommand{\val}[1]{\ensuremath{\mathit{#1}}}

\newcommand{\decp}[1]{\texttt{#1}}

\newcommand{\claim}{\textsc{Claim Mgmt}\xspace}
\newcommand{\bpicEleven}{\textsc{BPIC11}\xspace}
\newcommand{\bpicFifteen}{\textsc{BPIC15}}
\newcommand{\production}{\textsc{Production}\xspace}
\newcommand{\sepsis}{\textsc{Sepsis}\xspace}

\newlength{\maxlen}

\begin{document}
%
\title{Explain, Adapt and Retrain: How to improve the accuracy of a PPM classifier through different explanation styles}

%
%
%
%

\author{Williams~Rizzi,
        Chiara~Di Francescomarino,
        Chiara~Ghidini,
        and~Fabrizio Maria~Maggi.
\IEEEcompsocitemizethanks{
    \IEEEcompsocthanksitem W. Rizzi, C. Ghidini are with the PDI Unit, Fondazione Bruno Kessler, Trento, Italy. E-mail: wrizzi@fbk.eu, ghidini@fbk.eu \protect 
    \IEEEcompsocthanksitem C. Di Francescomarino is with the DISI Department, University of Trento, Trento, Italy. E-mail: c.difrancescomarino@unitn.it \protect 
    \IEEEcompsocthanksitem W. Rizzi, F. M. Maggi are with the KRDB Unit, Free University of Bozen Bolzano, Bolzano, Italy. E-mail: wrizzi@unibz.it, maggi@inf.unibz.it \protect 
}
\thanks{Manuscript received MONTH DAY, YEAR; revised MONTH DAY, YEAR.}}

%
%

\markboth{Journal of \LaTeX\ Class Files,~Vol.~14, No.~8, August~2015}%
{Shell \MakeLowercase{\textit{et al.}}: Bare Advanced Demo of IEEEtran.cls for IEEE Computer Society Journals}
%



\IEEEtitleabstractindextext{%
\begin{abstract}
    Recent papers have introduced a novel approach to explain why a Predictive Process Monitoring (PPM) model for outcome-oriented predictions provides wrong predictions. Moreover, they have shown how to exploit the explanations, obtained using state-of-the art post-hoc explainers, to identify the most common features that induce a predictor to make mistakes in a semi-automated way, and, in turn, to reduce the impact of those features and increase the accuracy of the predictive model. This work starts from the assumption that frequent control flow patterns in event logs may represent important features that characterize, and therefore explain, a certain prediction. Therefore, in this paper, we (i) employ a novel encoding able to leverage \declare constraints in \ppm and compare the effectiveness of this encoding with \ppm state-of-the art encodings, in particular for the task of outcome-oriented predictions; (ii) introduce a completely automated pipeline for the identification of the most common features inducing a predictor to make mistakes; and (iii) show the effectiveness of the proposed pipeline in increasing the accuracy of the predictive model by validating it on different real-life datasets.
\end{abstract}

}

\maketitle

\IEEEdisplaynontitleabstractindextext

%
\IEEEpeerreviewmaketitle

\ifCLASSOPTIONcompsoc
\IEEEraisesectionheading{\section{Introduction}\label{sec:introduction}}
\else
\section{Introduction}
\label{sec:introduction}
\fi

\IEEEPARstart{P}{redictive} (business) Process Monitoring is a family of techniques that use event logs to make predictions about the future state of the executions of a business process~\cite{DBLP:conf/caise/MaggiFDG14,DBLP:conf/bpm/DiFrancescomarino18}. For example, a \ppm technique may seek to predict the remaining execution time of each ongoing case of a process~\cite{DBLP:journals/tist/VerenichDRMT19}, the next activity that will be executed in each case~\cite{DBLP:conf/bpm/EvermannRF16}, or the final outcome of a case with respect to a set of possible business outcomes~\cite{DBLP:journals/tkdd/TeinemaaDRM19}. 

Together with \ppm approaches, several works have been recently focusing on explaining predictive models or why a certain prediction has been returned by a predictor. All these works aim at providing explanations to \ppm users so as to increase their trust in predictions.

A very recent approach~\cite{DBLP:conf/bpm/RizziFM20}, differently from previous ones, directed the efforts of producing explanations in \ppm towards the goal of increasing the accuracy of the predictor. The approach leverages post-hoc explainers for identifying and selecting -- with some user support -- the most common features that induce a predictor to make mistakes and aims at reducing the impact of these features in order to increase the accuracy of the predictive model. State-of-the-art \ppm encodings are used to train the classifier.
This work starts from the assumption that frequent control flow patterns, holding a key role in process mining, may be used as features for characterizing, and therefore explaining (incorrect) predictions. 

Therefore, in this paper, an encoding based on temporal declarative patterns is used for the first time for making predictions on the outcome of ongoing executions, as well as for explaining incorrect predictions. Moreover, a novel and completely automated pipeline is proposed for the identification of the features inducing a predictor making mistakes. The pipeline, as the one introduced in~\cite{DBLP:conf/bpm/RizziFM20}, leverages standard post-hoc explainers to understand which of the features characterize incorrect predictions and to retrain the predictive model by randomizing the values of these features in order to increase the accuracy of the predictor. 
To this aim, the proposed pipeline is based on three phases: a new phase (phase 2) for the selection of the best explanations for characterizing the incorrect predictions is introduced after the extraction of the frequent explanations (phase 1) and before applying feature shuffling and model re-training (phase 3).

The rest of the paper is structured as follows.
Section~\ref{sec:background} presents some background notions.
Section~\ref{sec:problem} introduces the problem we want to address, while Section~\ref{sec:declareEncoding} and Section~\ref{sec:approach} present, respectively, the novel encoding based on temporal declarative patterns and the novel pipeline we propose to solve the problem.
Section~\ref{sec:evaluation} and Section~\ref{sec:results} discuss, respectively, the evaluation of the approach and its results while Section~\ref{sec:related} summarizes the related work. Finally, Section~\ref{sec:conclusions} concludes the paper and spells out directions for future work.

\section{Background}
\label{sec:background}

In this section, we introduce some background notions on explainable machine learning techniques, event logs, Predictive Process Monitoring and state-of-the-art encodings in this field, as well as on the \declare language.

\subsection{Explainable machine learning techniques}
White-box machine learning algorithms like decision trees can be easily explained, e.g., by following the tree path which led to a prediction. However, predictions based on more complex algorithms are often incomprehensible by human intuition and, without understanding the rationale behind the predictions, users may simply not trust them. Explainability is motivated by this lack of transparency of black-box approaches, which do not foster trust and acceptance of machine learning algorithms. Explainers help addressing this problem by explaining predictions provided by black-box predictive models.

An explainer allowing its incorporation at testing or runtime is SHapley Additive exPlanations (SHAP)~\cite{DBLP:conf/nips/LundbergL17}. SHAP is a game-theoretic approach to explain the output of any machine learning model. It connects optimal credit allocation with local explanations using the classic Shapley values from game theory and their related extensions. SHAP provides local explanations based on the outcome of other explainers and representing the only possible consistent and locally accurate additive feature attribution method based on expectations.

\subsection{Event logs}
\label{ssec:event_log}
Event logs record the execution of business processes, i.e., \emph{traces}, and traces consist of events.
Each event in a log refers to an \emph{activity} (i.e., a well-defined step in a business process) and is related to a particular trace.
Events that belong to a trace  are \emph{ordered} and constitute a single ``run'' of the process.
For example, in trace $\textbf{t}_i=\left\langle event_1, event_2, \ldots event_n \right\rangle$, 
the first activity to be executed is the activity associated to $event_1$.

Events may be characterized by multiple \emph{attributes}, e.g., an event refers to an \emph{activity}, may have a \emph{timestamp}, may be executed or initiated by a given \emph{resource}, and may have associated other \emph{data attributes}, i.e., data produced by the activities of the process. 
We indicate the value of an attribute $a$ for an event $e$ with $\pi_{\act{a}}(e)$. Standard attributes are, for instance, $\pi_{\act{activity}}(e)$ representing the activity associated to event $e$; $\pi_{\act{time}}(e)$ representing the timestamp associated to $e$; $\pi_{\act{resource}}(e)$ representing the resource associated to $e$. Graphically, we denote traces with \emph{data attributes} associated to the events as 
\begin{align*}
	\begin{footnotesize}
	\text{(event$_1$\{associated data\}, $\dotsc$, event$_n$\{associated data\})}
	\end{footnotesize}
\end{align*}
Traces can also have attributes. Trace attributes, differently from event attributes, are not associated to a single event but to the whole trace. Data associated to events and traces in event logs are also called \emph{data payloads}.

\subsection{Predictive Process Monitoring}
\label{sub:ppm}

\ppm~\cite{DBLP:conf/caise/MaggiFDG14} is a branch of Process Mining that aims at predicting at runtime and as early as possible the future development of ongoing cases of a process given their uncompleted traces. In the last few years, a wide literature about \ppm techniques has become available - see~\cite{DBLP:conf/bpm/DiFrancescomarino18} for a survey - mostly based on machine learning techniques. The main dimension that is typically used to classify \ppm techniques is the type of prediction, which is usually classified in three macro-categories: numeric predictions (e.g., time or cost predictions); categorical predictions (e.g., risk predictions or specific categorical outcome predictions such as the fulfilling or not of a certain property); next activities predictions (e.g, the sequence of the future activities, possibly with their payloads). In this paper, we focus on categorical predictions. 

Frameworks such as \nirdi~\cite{DBLP:conf/bpm/RizziSFGKM19} collect a set of machine learning techniques that can be instantiated and used for providing different types of predictions to the user.
In particular, these frameworks take as input a set of past executions and use them to train predictive models that can then be stored to be used at runtime to continuously supply predictions to the user.
Moreover, the computed predictions can be used to compute accuracy scores for the trained models.
By looking at these frameworks, we can identify two main modules: one for the \emph{case encoding}, and one for the \emph{supervised classification learning}. Each of them can be instantiated with different techniques. In this paper, we focus on traditional machine learning approaches, and, in particular, on Random Forest as a supervised classification learning technique. Instead, we rely on two state-of-the-art case encodings, illustrated in Section~\ref{sub:encodings}, and we introduce a novel encoding based on the \declare language. 

\subsection{Trace Encodings}
\label{sub:encodings}

The goal of the Predictive business Process Monitoring techniques used in this paper is to build a classifier that learns from a set of historical traces $L$ how to discriminate classes of traces and predict as early as possible the outcome of a new, unlabeled trace. More specifically, we are interested in automatically deriving a function $f$ that, given an ongoing sequence $\sigma$ provides a label for it, i.e., \(f : (L, \sigma) \rightarrow \{label\}\). To achieve this goal, a classifier is trained on all sequence prefixes of the same length derived from labeled historical traces in $L$, where the \textit{label} represents the (binary or categorical) outcome, and is usually placed at the end of the sequence as follows:
\begin{align*}
	\begin{footnotesize}
	\text{(event$_1$\{associated data\}, $\dotsc$, event$_n$\{associated data\}): label}
	\end{footnotesize}
\end{align*}

Sequence prefixes are transformed into the numeric information a machine learning algorithm can use to train a classifier by means of encodings. More specifically, each (prefix) sequence $\sigma_i$, $i = 1...k$ has to be represented through a feature vector \(g_i=(g_{i1},g_{i2},...g_{ih})\).
In this paper, we exploit the \emph{simple-index} and \emph{complex-index}~\cite{DBLP:conf/bpm/LeontjevaCFDM15} encodings that we illustrate by means of the following sample traces:  
\begin{gather}
	\label{toyexample-1}
	 \resizebox{.7\hsize}{!}{\tag{$\sigma_1$}\text{(\act{Register}\{\val{33}, \val{financial}\},$\dotsc$,\act{Accept\ Claim}\{\val{33}, \val{assessment}\},$\dotsc$):true}}\\
	  \notag \vdots\\
	 \label{toyexample-k}
	 \resizebox{.75\hsize}{!}{\tag{$\sigma_k$} \text{(\act{Register}\{\val{56}, \val{financial}\},$\dotsc$, \act{Send\ Questionnaire}\{\val{56}, \val{secretary}\},$\dotsc$):false}}
\end{gather}
In the example, \act{Register} is the first event of sequence \eqref{toyexample-1}. Its data payload $\{ \val{33}, \val{financial} \}$ corresponds to the data associated to attributes \att{age} and \att{department}. Note that the value of \att{age} is static: it is the same for all the events in a trace (trace attribute), while the value of \att{department} is different for every event (event attribute).
In case for some event the value for a specific attribute is not available, the value \emph{unknown} is specified for it.

\begin{table}[t]
\centering
\caption{Simple-index encoding.}
\scalebox{0.9}{%
\begin{tabular}{lcccccc}
	\toprule
	& event\_1  & $\ldots$ & event\_i    & $\ldots$& event\_m  & label \\ 
	\midrule
	$\sigma_1$         & \act{Register} &  & \act{Accept\ Claim} &    & \act{Archive} &     true \\
	& & & $\ldots$                       \\
	$\sigma_k$        & \act{Register}  &  & \act{Send\ Questionnaire}  &    & \act{Archive}    & false  \\ 
	\bottomrule
\end{tabular}			
}
\label{tab:simple}
\end{table}

\begin{table*}[t]
\centering
\caption{Complex-index encoding.}
\scalebox{0.58}{
\begin{tabular}{lccccccccccccc} 
	\toprule
	& \att{age} & event\_1 & $\ldots$ & event\_i    & $\ldots$ & event\_m   & $\ldots$ & \att{department\_1} & $\ldots$ & \att{department\_i} & $\ldots$ & \att{department\_m} & label \\ 
	\midrule
	\multirow{2}{*}{$\sigma_1$} & \multirow{2}{*}{\val{33}} & \multirow{2}{*}{\act{Register}} &       & \act{Accept} &  & \multirow{2}{*}{\act{Archive}} &  & \multirow{2}{*}{\val{financial}}  &  & \val{assessment} & & \multirow{2}{*}{\act{Archive}} & \multirow{2}{*}{true} \\
															& 										& 																&       & \act{Claim}  &  &															 &  & 															 &  & \val{dept} 			 & & 																& 												\\
	&&& $\ldots$      	\\
	\multirow{2}{*}{$\sigma_j$}  & \multirow{2}{*}{\val{56}} & \multirow{2}{*}{\act{Register}} &     & \act{Send}    					& & \multirow{2}{*}{\act{Archive}} &    & \multirow{2}{*}{\val{financial}}   &     & \multirow{2}{*}{\val{secretary}}  & & \multirow{2}{*}{\act{Archive}} & \multirow{2}{*}{false}  \\ 
															 & 										 & 																 &     & \act{Questionnaire}    & &                       &    &    						 &     &   													 & & 															  & \\ 	
	\bottomrule
\end{tabular}
}
\label{tab:complex}
\end{table*}

The \emph{simple-index} encoding takes into account the control flow of a trace, and, in particular, information about the order in which events occur in the sequence. Here, each feature corresponds to a position in the sequence and the possible values for each feature are the activities. By using this type of encoding, traces \eqref{toyexample-1} and \eqref{toyexample-k} would be encoded as reported in Table~\ref{tab:simple}.

The \emph{complex-index} encoding takes into account both the control flow and the data payloads of a trace. In particular, the data associated with events in a sequence is divided into static and dynamic information. Static information is the same for all the events in the sequence (e.g., the information contained in trace attributes), while dynamic information changes for different events (e.g., the information contained in event attributes).
The resulting feature vector $g_i$, for a sequence $\sigma_i$, is: 
\begin{align*}
	g_i = (& s^1_{i},\dotsc,s^u_{i}, event_{i1}, event_{i2}, \dotsc, event_{im}, \\
		& h^{1}_{i1}, h^{1}_{i2},\dotsc,h^{1}_{im},\dotsc,h^{r}_{i1}, h^{r}_{i2},\dotsc,h^{r}_{im}),
\end{align*}
where each $s_{i}$ is a static feature, each $event_{ij}$ is the activity at position j and each $h_{ij}$ is a dynamic feature associated to an event. For example, by using this type of encoding, traces \eqref{toyexample-1} and \eqref{toyexample-k} would be encoded as shown in Table~\ref{tab:complex}.

\subsection{\declare constraints}
\label{sub:declare}
\declare is a declarative process modeling language originally introduced by Pesic and van der Aalst in \cite{Pesic2007:DECLARE}. \declare constraints are temporal properties expressed using rules that specify possible orderings of activities (see \tablename~\ref{tbl:timed-ltl}).

\begin{table}[tb]
	\caption{\declare constraints and their semantics\label{tbl:timed-ltl}.}
	\centering
	\scalebox{0.9}{
		\begin{tabular}{ll}
			\toprule
			\textbf{Constraint}    & \textbf{Semantics} \\
			\midrule
			\decp{existence} (\act{a}) &  \act{a} always eventually occurs \\
			\decp{response} (\act{a},\act{b}) &  \act{a} is always eventually followed by \act{b} \\
			\decp{chain response} (\act{a},\act{b}) &  \act{a} is always immediately followed by \act{b} \\
			\decp{precedence} (\act{a},\act{b}) &  \act{b} is always preceded by \act{a}\\
			\decp{not succession} (\act{a},\act{b}) & \act{a} is never eventually followed by \act{b}\\
			\decp{coexistence} (\act{a},\act{b})  & \act{a} and \act{b} can only occur together \\
			\bottomrule
		\end{tabular}
	}
\end{table}
Consider, for example, the \emph{response} constraint indicating that if activity \act{a} {\it occurs}, activity \act{b} must
eventually {\it follow}. This constraint is satisfied for traces such as $\textbf{t}_1$ = $\langle \act{a}, \act{a},\act{b}, \act{c} \rangle$, $\textbf{t}_2 = \langle \act{b},
\act{b}, \act{c}, \act{d} \rangle$, and $\textbf{t}_3 = \langle \act{a}, \act{b}, \act{c}, \act{b} \rangle$, but not for $\textbf{t}_4 = \langle \act{a}, \act{b}, \act{a}, \act{c} \rangle$ because, in this case, the second instance of  \act{a} is not followed by a \act{b}. Note that, in $\textbf{t}_2$, the considered \decp{response} constraint is satisfied in a trivial way because \act{a} never occurs. In this case, we say that the constraint is \emph{vacuously satisfied}~\cite{kupf:vacu03}.

An \emph{activation} of a constraint in a trace is an event whose
occurrence imposes, because of that constraint, some obligations on other events (targets)
in the same trace. For example, \act{a} is an activation for the considered \decp{response}
constraint and \act{b} is a target, because the execution of \act{a} forces \act{b} to be executed, eventually.

An activation of a constraint can be a \emph{fulfillment} or a \emph{violation}
for that constraint. When a trace is perfectly compliant with respect to a constraint,
every activation of the constraint in the trace leads to a fulfillment.
Consider, again, the \decp{response} constraint indicating that if \act{a} {\it occurs}, \act{b} must
eventually {\it follow}. In trace $\textbf{t}_1$, the
constraint is activated and fulfilled twice, whereas, in trace $\textbf{t}_3$,
the same constraint is activated and fulfilled only once. On the other hand,
when a trace is not compliant with respect to a constraint, an activation of the
constraint in the trace can lead to a fulfillment but also to a violation (at
least one activation leads to a violation). In trace $\textbf{t}_4$, for
example, the \decp{response} constraint is activated twice, but the first activation leads to a
fulfillment (eventually \act{b} occurs) and the second activation
leads to a violation (\act{b} does not occur subsequently).

\begin{figure*}[h!]
	\centering
		\includegraphics[width=.9\textwidth]{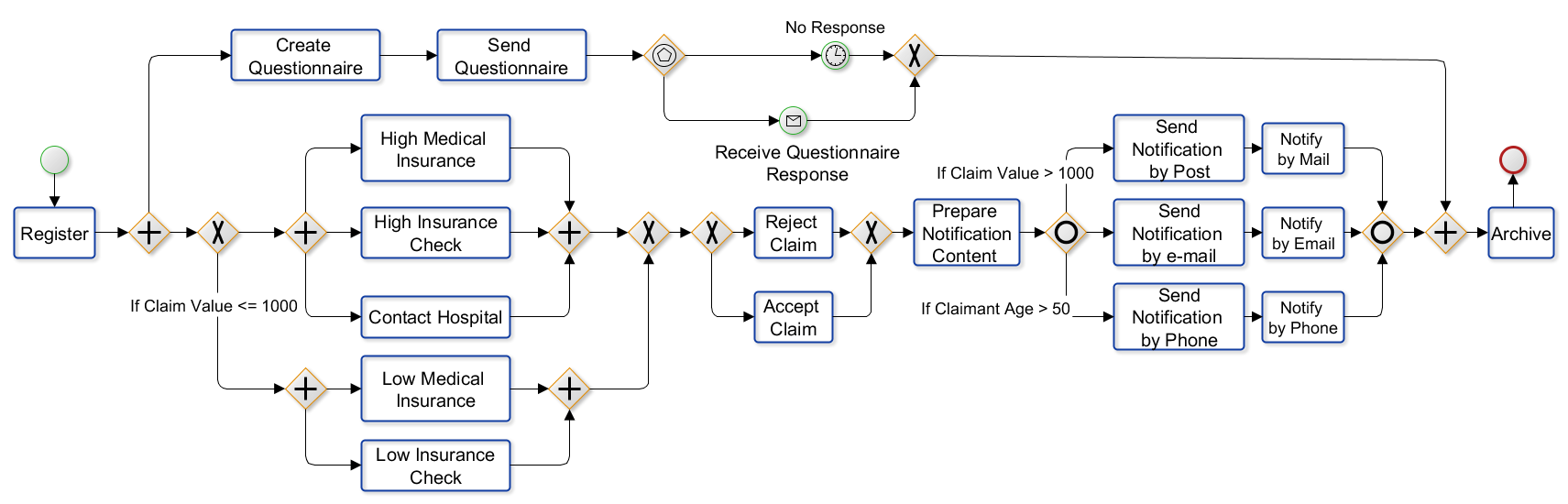}
	\caption{Running example.}
	\label{fig:driftExample}
\end{figure*}

\section{Problem}
\label{sec:problem}

To illustrate the problem we want to address in this paper, we use the process model introduced in~\cite{DBLP:conf/IEEEscc/MaisenbacherW17} that we report in Figure~\ref{fig:driftExample} using the BPMN language.
The process considers two different checks of claims, a basic one and a complex one, depending on the value of the claim. After the decision is taken about the claim, the claimant is notified using different communication means depending on some specific conditions.
In parallel, a questionnaire is sent to the claimant which can be received back before a certain deadline.

We can consider a labeling that classifies process executions in \emph{accepted claims} (positive) and \emph{rejected claims} (negative) based on whether the trace contains a claim acceptance (\act{Accept\ Claim}) or a claim rejection (\act{Reject\ Claim}). It is well known that the predictions provided by the classifier about the outcome of an ongoing process execution are not always correct and the accuracy of the classifier is not always optimal. This is due to the fact that, to make a prediction, the wrong information could be taken into consideration by the classifier. This is because the training set can contain noise or because the learning algorithm could not discover the ``perfect'' classifier (for example, if the data provided in the training set is not sufficient to identify the correct correlation function between feature vectors and labels, or if it contains noise).

We can assume, for instance, that the dataset used to train the predictive model to make predictions at runtime about the acceptance or rejection of a claim is noisy and that the classifier learns that a process execution is positive also in those cases in which after a claim is rejected the questionnaire is sent. If this correlation is learnt by the predictive model, when, at runtime, a questionnaire is sent after the claim has been rejected the predictive model will return a false positive.

Using an explainer such as SHAP~\cite{DBLP:conf/nips/LundbergL17}, the user can identify the features that have influenced the most the classifier when giving a wrong prediction. In this respect, our overall objective is that of taking into account the features that have the highest influence on the classifier when it provides the wrong predictions according to the explainer, so as to to improve the accuracy of the classifier by reducing the impact of these features. To this aim, we retrain the classifier after randomizing the values of all the features that are important when the predictor fails, but have low or no importance when it provides the correct predictions.

In our example, let us assume that a wrong prediction occurs when a claim is rejected and a questionnaire is sent afterwards. Our goal would be to be able to randomize the occurrence of \act{Reject\ Claim} eventually followed by \act{Send\ Questionnaire} and then retrain the classifier. When trying to identify the features that characterize (explain) a prediction (a classification in our example), an explainer such as SHAP is tight to the way features are represented in the encoding of the trace. Unfortunately, no encodings exist that would enable the identification of the occurrence of \act{Reject\ Claim} eventually followed by \act{Send\ Questionnaire} in an arbitrary position of the trace and with an arbitrary distance between the two activities, and more in general the representation of temporal control flow patterns occurring in an event log. 
For instance, using the simple index encoding one would discover all the specific occurrences of \act{Reject\ Claim} at position $i$ followed by the specific occurrences of \act{Send\ Questionnaire} at position $j$ (for all $i,j$ within the trace length). This may lead to a vague and sparse representation of our concise pattern in several instances and may lead to the inability of identifying it. This problem can be solved by following the commonly recognized view that frequent temporal patterns are sometimes crucial to support process analysis and that, in some scenarios, it is convenient to describe traces according to them via declarative modeling languages such as \declare.    

Therefore, our additional objective here is that of enabling explainers to discover features in terms of \declare constraints and to investigate their effectiveness when re-training a predictive model.

\section{\declare Encoding}
\label{sec:declareEncoding}

In order to enable explainers to discover features in terms of \declare constraints, we leverage, for the first time 
in Predictive Process Monitoring, 
the \declare encoding introduced in~\cite{10.48550/arxiv.2111.12454} for the discovery of deviant traces. In the \declare encoding, each element of the feature vector corresponds to a grounded \declare constraint (taken from a list of selected temporal patterns) and has value:

\begin{itemize}
\item -1, if the corresponding \declare constraint is violated in the trace;
\item 0, if the corresponding \declare constraint is vacuously satisfied in the trace;
\item $n$, if the corresponding \declare constraint is satisfied and activated $n$ times in the trace.
\end{itemize}
The event log is, therefore, transformed into a matrix of numerical values where each row corresponds to a trace and each column corresponds to a \declare constraint. For instance, given trace $\left\langle \act{a},\act{b},\act{c},\act{a},\act{b},\act{c},\act{d},\act{a},\act{b}\right\rangle$:
\begin{itemize}
\item constraint \decp{response}(\act{a},\act{c}) is violated, since the third activation (the third occurrence of \act{a}) leads to a violation (is not eventually followed by \act{c}) and is encoded as -1;
\item constraint \decp{response}(\act{a},\act{b}) is satisfied and activated 3 times and is encoded as 3;
\item constraint \decp{response}(\act{d},\act{b}) is vacuously satisfied and is encoded as 0.
\end{itemize} 

To select the \declare constraints to be used for building the feature vectors, we first 
 discover a list of constraints from the event log (specifically, from the training set). To this aim, we rely on the initial steps of the algorithms for the discovery of \declare models~\cite{MaCV12}. We discover frequent activity sets, that is sets of highly correlated activities. Highly correlated sets are then used to build, in any possible ways, the \declare constraints. For example, considering the frequent activity set $\{ \act{a}, \act{b} \}$, and the constraint type \decp{response}, the two \declare constraints \decp{response}(\act{a},\act{b}) and \decp{response}(\act{b},\act{a}) are generated. Each candidate is finally checked over the input log to verify if it is satisfied in a percentage of traces that is above a given \emph{support} threshold.

Finally, subsumed \declare constraints are filtered out so as to keep only the strongest ones. The remaining constraints are used as features in the \declare encoding.

\section{Approach}
\label{sec:approach}

\begin{figure*}[t]
	\centering
		\includegraphics[width=\textwidth]{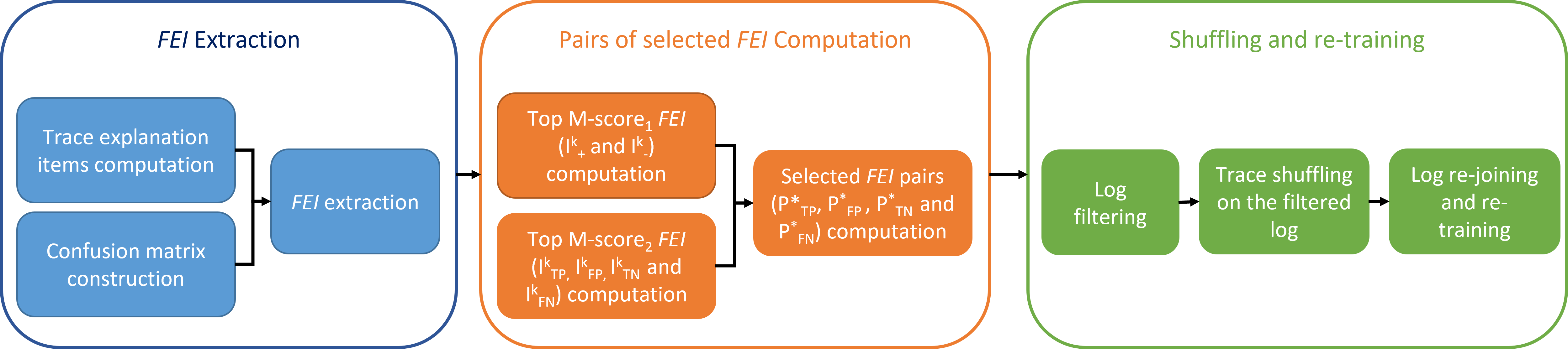}
	\caption{Approach}
	\label{fig:pipeline}
\end{figure*}

In this section, we describe the approach for explaining why a binary\footnote{In this paper, the proposed approach is applied to binary predictions. However, the approach can be generalized to multi-class predictions (see Section~\ref{sec:conclusions}).} outcome-based predictive model is wrong and eventually improving its performance. The proposed approach leverages a labeled event log, part of which is used for training a predictive model (\emph{training set}) and optimizing the hyperparameters (\emph{validation set}), while another part is used for identifying features and values characterizing wrong predictions (\emph{feedback set}). 

The approach, depicted in \figurename~\ref{fig:pipeline}, is based on three main steps. (i) First, a predictive model is trained and the \feis (\FEIs) are extracted (Section~\ref{ssec:extraction}). \FEIs are sets of frequent feature-value pairs explaining correct and wrong predictions on the feedback set. (ii) The most important \FEIs characterizing correct and wrong predictions are then selected (Section~\ref{ssec:selection}). (iii) Finally, the most important \FEIs are leveraged in order to shuffle the original training set in order to improve the performance of the predictive model by ``correcting'' the wrong correlations learned by the model (Section~\ref{ssec:shuffling}). The  proposed pipeline is a novel version of the approach reported in~\cite{DBLP:conf/bpm/RizziFM20}. It enhances and automates the original one by introducing a sophisticated mechanism for the automatic selection of the most important \FEIs -- see step (ii).

\subsection{\fei extraction}
\label{ssec:extraction}
In order to extract a characterization of the wrong predictions returned by the predictive model on the feedback set, we leverage the explanations of post-hoc model-agnostic explainers, such as SHAP~\cite{DBLP:conf/nips/LundbergL17}. This type of explainers returns an explanation for each prediction, in terms of the impact of each feature (and corresponding value) towards the prediction. 
For each trace (prefix) in the feedback set, we hence apply SHAP to explain the corresponding predictions. 
Then, in order to identify the explanations related to the wrong predictions, we compute the confusion matrix on the feedback set. We finally use the obtained explanations in order to characterize each quadrant of the confusion matrix by identifying sets of frequent feature-value pairs for traces in that specific quadrant (\FEIs). We detail in the following these three steps reported also in the first block of \figurename~\ref{fig:pipeline}.

\textbf{Computing explanation items for each trace prefix.}
The explainer takes as input the trained predictive model and the trace prefix whose outcome we want to predict, encoded according to one of the encodings (e.g., \simple, \complex or \declare), and returns as output, for each feature and corresponding value, an importance score value indicating the impact of the feature-value pair towards the prediction. The value of the importance score ranges between -1 and 1. A positive value means that the feature-value pair contributes to push the classifier towards the current prediction, while a negative value means that the feature-value pair contributes to push the classifier towards a different prediction. 

The absolute value of the score denotes the strength of the impact of the feature-value pair on the prediction.

For instance, we can consider a trace prefix in the feedback set of the claim management example described in Section~\ref{sec:problem} for which the predictive model has returned a positive prediction. Let us assume that the trace attribute \att{CType}, which specifies whether the customer making the claim is a \val{regular}, \val{silver}, \val{VIP} or \val{gold} customer, is included among the features of the trace encoding (e.g., by using the \complex encoding), and that the value of the feature is \val{Gold} for the specific trace. If the importance score returned by the explainer for the feature-value pair (\att{CType},\val{Gold}) is a positive value, the feature-value pair has contributed towards a positive prediction and the score indicates the strength of the contribution.

The feature-value pairs and the corresponding importance score returned by the explainer for each trace prefix (\emph{explanation items}) are then filtered out based on a user-defined threshold $t$, so as to keep only the feature-value pairs impacting the most on the prediction.

\begin{figure}[tb]
	\centering
		\includegraphics[width=.48\textwidth]{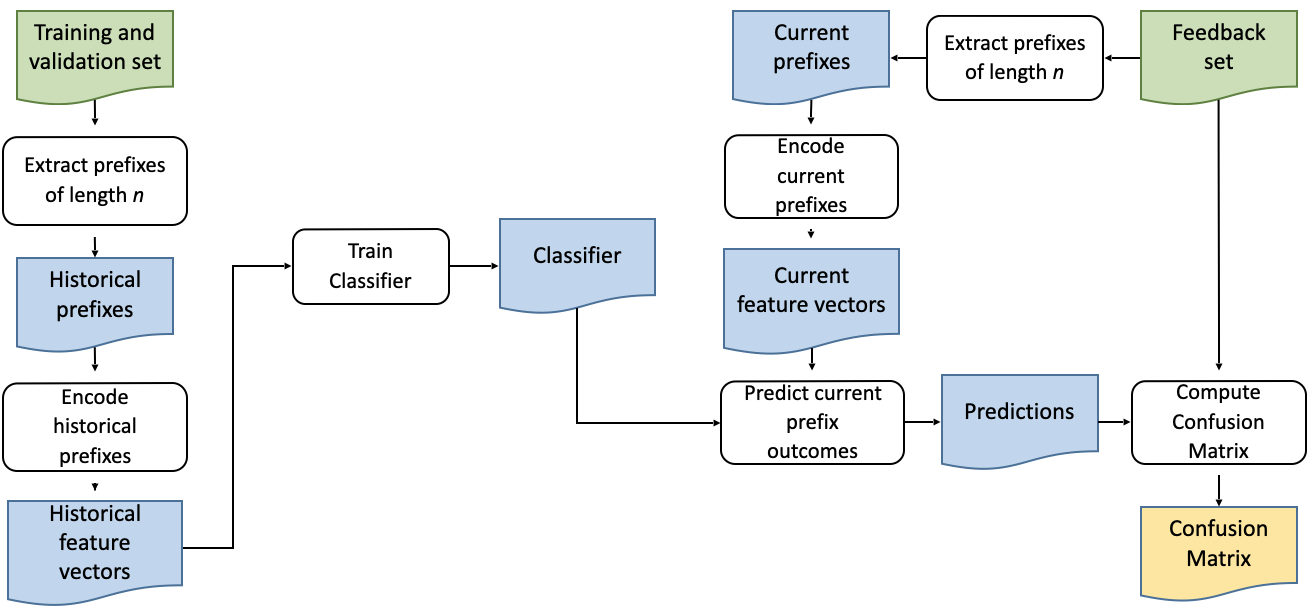}
	\caption{Confusion matrix construction.}
	\label{fig:confusion}
\end{figure}

\textbf{Building the confusion matrix.}
\figurename~\ref{fig:confusion} shows the approach we use to extract the confusion matrix from the training, validation and feedback set. The prefixes of length $n$ are extracted from the training and validation set and encoded with a specific encoding type. The encoded traces are then used to train a classifier and optimize its hyperparameters.  Prefixes of length $n$ are extracted also from the feedback set and used to query the classifier that returns the label that is the most likely outcome for the current prefix according to the information derived from the historical prefixes. The predicted labels are compared with the labels available in the feedback set (the \emph{gold standard}) and the \emph{confusion matrix} is built. 
The confusion matrix classifies predictions in four categories:
\begin{enumerate*}[(i)]
\item true-positive ($T_P$: positive outcomes correctly predicted);
\item false-positive ($F_P$: negative outcomes predicted as positive);
\item true-negative ($T_N$: negative outcomes correctly predicted);
\item false-negative ($F_N$: positive outcomes predicted as negative).
\end{enumerate*}

\textbf{Aggregating single trace explanation items.}
Explainers provide sets of feature-value pairs (and impact score) separately for each trace. From the previous steps, we hence get, for each quadrant of the confusion matrix, a set of top \emph{explanation items} for each feedback trace in that quadrant of the confusion matrix. 

With the aim of characterizing each quadrant as a whole, we extract association rules~\cite{10.1145/170036.170072} for each quadrant starting from the top \emph{explanation items} separately discovered for each trace. In particular, for each quadrant of the confusion matrix, we mine a set of frequent \FEIs. Each \FEI in a quadrant represents a conjunction of feature-value pairs that characterize (part of) the traces in that quadrant. The disjunction of the \FEIs in a quadrant provides a characterization of most of the traces in the quadrant. For instance, for the $T_P$ quadrant, we extract a set of \FEIs $I_{T_P}$ containing the frequent explanation items characterizing the traces for which a correct positive prediction has been returned. This step allows us to aggregate single trace explanations into explanations related to multiple traces. For example, if (\att{CType},\val{Gold}) is among the top explanation items of most of the traces with a positive prediction and a positive ground-truth outcome, \{(\att{CType},\val{Gold})\} is one of the \FEI in $I_{T_P}$ providing a characterization of the $T_P$ quadrant of the confusion matrix. This means that the traces with \att{CType} set to \val{Gold} are, in most of the cases, correctly predicted as positive by the predictive model.

\subsection{\fei selection}
\label{ssec:selection}

In order to break the wrong correlations learned by the model but not the correct ones in the shuffling step, we aim at perturbating only the features related to the traces that have been incorrectly classified. Among the several \FEIs generated from the extraction step, we hence want to select the ones characterizing the traces in the ``wrong'' quadrants (false positive and negative quadrants) but not the ones in the ``correct'' quadrants (true positive and negative quadrants). We therefore want to identify the \FEIs that best discriminate among quadrants (see the second block in \figurename~\ref{fig:pipeline}). 

To this aim, we introduce the \mscore, a score measuring the discriminativeness of a \FEI $i$ in characterizing a class $cl$ (with respect to a class $\neg cl$) on a set of traces $T$. Intuitively, the \mscore computes the difference between the percentage of traces characterized by the items in $i$ (i.e., satisfying all the items in $i$) and belonging to the class $cl$ and the percentage of traces characterized by the items in $i$ and belonging to the class $\neg cl$. Given the subset of traces belonging to the class $cl$ ($T^{cl} \subseteq T$), the subset of traces belonging to the class $\neg cl$ ($T^{\neg cl} \subseteq T$) and the subset of traces satisfying all the items in the \FEI $i$ ($T^{i} \subseteq T$), the \mscore is computed as reported in Equation~\ref{maggi_score}:
\begin{equation}
		M_{(cl,T)}(i) = \frac{\lvert T^{cl} \cap T^i \rvert}{\lvert T^{cl} \rvert} - \frac{\lvert T^{\neg cl} \cap T^{i} \rvert}{\lvert T^{\neg cl} \rvert}.
	\label{maggi_score}
\end{equation} 

For instance, given the class of the positive predictions $+$ and the class of the negative predictions $-$, the \mscore $M_{(+,T)}(i)$ of a \FEI $i$ will tell us how well $i$ discriminates between positive and negative predictions.

As reported in the second block in \figurename~\ref{fig:pipeline}, in order to identify the most discriminative \FEIs for each quadrant, we use the \mscore for discriminating (i) between positive and negative predictions ($M$-score$_{1}$); (ii) between correct and wrong predictions ($M$-score$_{2}$).
First, we use the \mscore metrics to identify the \FEIs related to the quadrants with a positive prediction (negative prediction) that best discriminate between the class of the positive versus negative (negative versus positive) predictions on the whole set of traces. This means that we compute $M_{(+,T)}$, for the \FEIs in $I_{+}=I_{T_P} \cup I_{F_P}$, and we compute $M_{(-,T)}$, for the \FEIs in $I_{-}=I_{T_N} \cup I_{F_N}$.
We then use the \mscore to identify the \FEIs that best discriminate between the class of the correct versus wrong predictions on the set of traces for which the predictive model returns a positive prediction, as well as on the set of traces for which the predictive model returns a negative prediction. This means that for discriminating between correct and wrong positive predictions we compute $M_{(c,T^{+})}$ for the \FEIs in $I_{T_P}$ and we compute $M_{(w,T^{+})}$ for the \FEIs in $I_{F_P}$. Symmetrically, for discriminating between correct and wrong negative predictions, we compute $M_{(c,T^{-})}$, for the \FEIs in $I_{T_N}$, and $M_{(w,T^{-})}$, for the \FEIs in $I_{F_N}$.

By using the \mscore for ranking the \FEIs according to the two steps described above, we finally get six ordered sets of \FEIs: $I_{+}$ and $I_{-}$ ordered based on their capability to discriminate between positive and negative (negative and positive) predictions; $I_{T_P}$ and $I_{F_P}$ ranked based on their capability to discriminate between correct and wrong (wrong and correct) positive predictions; and $I_{T_N}$ and $I_{F_N}$ ranked based on their capability to discriminate between correct and wrong (wrong and correct) negative predictions.
For each of the six ranked sets we select the top $k$ \FEIs ($I^{k}_{x}$), given a user-defined threshold $k$. Intuitively, the $k$ selected \FEIs can be seen as a disjunction of \FEIs, each characterizing a subset of the traces that the itemsets in $x$ are able to discriminate. 

Starting from these six sets of \FEIs, we are able to identify the itemsets that characterize each of the four quadrants of the confusion matrix. We can then use these characterizations in order to identify the itemsets causing the predictive model to learn wrong correlations for the $F_P$ and $F_N$ quadrants.
To this aim, we combine each of the top \FEIs characterizing positive and negative predictions ($i_1 \in I^{k}_{+} \cup I^{k}_{-}$) with each of the top \FEIs characterizing correct and wrong predictions ($i_2 \in I^{k}_{T_P} \cup I^{k}_{F_P} \cup I^{k}_{T_N} \cup I^{k}_{F_N}$) into $k^2$ pairs of \FEIs $(i_1 \cup i_2,i)$, where $i=\emptyset$ in the case of correct prediction quadrants ($T_P$ and $T_N$) and $i=i_2$ in the case of incorrect prediction quadrants ($F_P$ and $F_N$). The first element of the pair represents a characterization of a subset of the traces contained in the specific quadrant. The second element of the pair, instead, represents the feature-value pairs that have to be shuffled to weaken the wrong correlations. Specifically, we have that:
\begin{itemize}
\item for the $T_P$ quadrant, we compute the set $P^*_{T_P}$, such that $\forall i_1 \in I^k_{+}$ and $\forall i_2 \in I^k_{T_P}$, $(i_1 \cup i_2,\emptyset) \in P^*_{T_P}$. The $T_P$ quadrant is hence described by the \FEIs $i_1 \cup i_2$, and no feature has to be shuffled as predictions in this quadrant are correct.
\item for the $F_P$ quadrant, we compute the set $P^*_{F_P}$, such that $\forall i_1 \in I^k_{+}$ and $\forall i_2 \in I^k_{F_P}$, $(i_1 \cup i_2,i_2) \in P^*_{F_P}$. The $F_P$ quadrant is hence described by the \FEIs $i_1 \cup i_2$, and the feature-value pairs in $i_2$, which specifically characterize the wrong predictions for this class of traces, are the ones to be shuffled.
\item for the $T_N$ quadrant, we compute the set $P^*_{T_N}$, such that $\forall i_1 \in I^k_{-}$ and $\forall i_2 \in I^k_{T_N}$, $(i_1 \cup i_2,\emptyset) \in P^*_{T_N}$. The $T_N$ quadrant is hence described by the \FEIs $i_1 \cup i_2$, and no feature has to be shuffled as predictions in this quadrant are correct.
\item for the $F_N$ quadrant, we compute the set $P^*_{F_N}$, such that $\forall i_1 \in I^k_{-}$ and $\forall i_2 \in I^k_{F_N}$, $(i_1 \cup i_2,i_2) \in P^*_{F_N}$. The $F_N$ quadrant is hence described by the \FEIs $i_1 \cup i_2$, and the feature-value pairs in $i_2$, which specifically characterize the wrong predictions for this class of traces, are the ones to be shuffled.
\end{itemize}

Note that the set of \FEIs characterizing each quadrant can be seen (in a more readable format) as a conjunction of formulas. For instance, if the set of \FEIs \{(\att{CType}, \val{Gold}), (\att{PClaims}, \val{No})\} characterizes the set of false positives, this means that the false positive traces are characterized by the formula \att{CType}=\val{Gold} and \att{PClaims}=\val{No}.

\subsection{Shuffling and re-training}
\label{ssec:shuffling}

The pairs of \FEIs extracted from the previous step can be used to weaken the incorrect correlations learned by the predictive model. To this aim, we shuffle, in the feature vectors representing the traces of the training (and the validation) set, the values of the features that have brought the predictive model to learn these correlations, and, then, we re-train the predictive model (last block in \figurename~\ref{fig:pipeline}). 

In order to weaken only the wrong correlations, while preserving the correct ones learned by the predictive model, for each pair of \FEIs $(ic,\emptyset)$ in $P^*_{T_P}$ and $P^*_{T_N}$, we first filter out from the training (and the validation) set the traces satisfying all the items in $ic$, that is the traces characterized by correlations ``correctly'' learned by the model ($T_{ic}$).
Among the remaining traces ($T \setminus \bigcup_{ic}{T_{ic}}$), for each pair of \FEIs $(iw_1,iw_2)$ in $P^{*}_{F_P}$ and $P^{*}_{F_N}$, the traces satisfying all the items in $iw_1$ are selected ($T_{iw_1}$), that is the traces characterized by correlations ``incorrectly'' learned by the model, and the feature-value pairs for each item in $iw_2$ are shuffled. 

The shuffling procedure differs according to the specific encoding used. In the case of the \simple and \complex encodings, for each pair of \FEIs $(iw_1,iw_2)$ in $P^{*}_{F_P}$ and $P^{*}_{F_N}$, and, for each item in the \FEI $iw_2$, the value of the feature in the trace $t \in T_{iw_1}$ is replaced by another one, randomly chosen among the admissible values for that feature. For instance, if the feature-value pair to be shuffled is $(\act{CType},\val{VIP})$, the value \val{VIP} is replaced by one of the admissible values for the feature \att{CType}, that is one among ${\val{Regular},\val{Silver},\val{Gold},\val{VIP}}$.

When the \declare encoding is used, the shuffling procedure is slightly different as changing the value of a feature also possibly requires changing the value of other features. 
For instance, consider trace $\act{t}=\langle \act{b} \rangle$ and its \declare encoding relying on only four \declare features: \decp{existence(\act{a})}, \decp{existence(\act{b})}, \decp{response(\act{a},\act{b})}, \decp{response(\act{b},\act{a})}, thus resulting in the encoded trace $\left[-1, 1, 0, -1\right]$. Let us assume that we want to shuffle the feature-value pair (\decp{existence}(\act{a}),-1) with the shuffled value 1. Changing only the first value of the encoded trace would not be enough as the occurrence of \act{a} also causes that constraint \decp{response(\act{a},\act{b})} becomes non-vacuously satisfied (it could be satisfied or violated depending on where \act{a} occurs). In order for the encoding to be consistent with the actual trace, we may hence need to change also the values of other features.  

In order to take into account such a dependency among the \declare features, we resort to aligning the whole trace to the shuffled value of the \declare constraint~\cite{DBLP:conf/aaai/XuLZ17a}. In particular, we align the trace corresponding to the encoded trace to the \declare constraint with the shuffled value and we then encode back the aligned trace. 

For instance, in the example above, we would align trace \act{t} to the \declare constraint \decp{existence}(\act{a})=1. The result of the alignment procedure will be, for instance, trace $\act{t'}=\langle \act{a}, \act{b} \rangle$. Encoding back the trace, using the four \declare features mentioned above, will result in the consistent shuffling of all the needed features: $\left[1, 1, 1, -1\right]$.

Once the shuffling procedure has been applied to all traces in $T_S = T \setminus \bigcup_{ic}{T_{ic}}$, and the new set $T^*_S$ with the shuffled traces has been computed, the new training (and validation) set $T^*=T^*_S \cup \bigcup_{ic}{T_{ic}}$ is (are) used to re-train the predictive model.

\section{Evaluation}
\label{sec:evaluation}

The evaluation reported in this paper aims at understanding how the different encodings affect the quality of the predictive model and the re-trained predictive model in terms of accuracy. 
In this section, we introduce the research questions, the datasets, the experimental setting and the metrics used to evaluate the effectiveness of the approach described in Section~\ref{sec:approach}. The results are instead reported in Section~\ref{sec:results}.

\subsection{Research Questions}
\label{ssec:research_questions}
Our evaluation is guided by the following research questions:

\begin{enumerate}[label=\textbf{RQ\arabic*.}, align=left]
	\item How do different \ppm encodings perform in terms of prediction accuracy?
	\item How effective is the re-training procedure in \ppm?
\end{enumerate}

\textbf{RQ1} aims at evaluating the quality of the predictions returned by the investigated encodings -- including the \declare one,

while \textbf{RQ2} investigates the quality of the predictions obtained by the retrained predictive models for the investigated \ppm encodings.

\subsection{Datasets}
\label{ssec:dataset}
We carried out the evaluation on both synthetic and real-life datasets. We used three synthetic datasets each specifically constructed for evaluating one of the three considered encodings. We instead evaluated all the three encodings against all the real-life datasets. 

Concerning the synthetic datasets, we considered the \claim event log 
derived from the process reported in Section~\ref{sec:problem}. It consists of $4800$ traces with an average trace length of $11$ events. The log contains $52\,935$ events and $16$ different activities. The three synthetic datasets were derived from this log by considering three different labelings, reported 

in the column \emph{Condition} of \tablename~\ref{tab:noise_conditions}. 
In addition, in order to evaluate the approach with respect to the three different considered encodings, we have systematically introduced some noise in the data used to train the classifier, so as to induce the classifier to make mistakes in a controlled way. In particular, we slightly changed the labeling of the training set using the conditions reported in the column \emph{Noise Condition} of \tablename~\ref{tab:noise_conditions}, so that the classifier will learn the noise condition rather than the proper condition. Each of the three labelings aims at specifically evaluating one of the three encodings (\simple, \complex and \declare, see column \emph{Encoding} in the table). Indeed, they are in a format that we expect to be easier to explain with the \simple, \complex and \declare encoding, respectively. For instance, the noise condition related to the \simple encoding will force the classifier to learn that only traces with \act{Accept Claim} in the fifth position will have a positive outcome. At evaluation time, this will result in several incorrect predictions: indeed, for all traces in which \act{Accept Claim} does not occur in position $5$, we will incorrectly get a negative outcome prediction (false negatives).

\setlength{\tabcolsep}{6pt}
\begin{table*}[h!]
	\centering
\scalebox{.7}{
\begin{tabular}{l l l l l}
\toprule
\textbf{Log} & \textbf{Labeling} & \textbf{Condition} & \textbf{Noise Condition}  & \textbf{Encoding}\\
\midrule
\multirow{6}{*}{\claim}  & $\phi_{S1}$                  & \decp{existence}(\act{Accept Claim})   & $\pi_{\act{act}}(e_5)$ = \act{Accept Claim}              & \simple \\ \cmidrule{2-5}
                         & \multirow{2}{*}{$\phi_{S2}$} & (\att{Age}$<60$) and                   & (\att{Age}$<60$) and                                     & \multirow{2}{*}{\complex} \\
                         &                              & (\att{CType}$=$\val{Gold})             & (\att{CType}$=$\val{Gold} or \att{CType}$=$\val{Silver}) & \\ \cmidrule{2-5}
                         & \multirow{2}{*}{$\phi_{S3}$} & \multirow{2}{*}{\decp{existence}(\act{AcceptClaim})} &  \decp{existence}(\act{AcceptClaim}) and                                 & \multirow{2}{*}{\declare}\\
                         &                              &                                                      & (!\decp{response(\act{Low Medical History},\act{Create Questionnaire})}) &                    \\

\bottomrule
\end{tabular}
}
\caption{Synthetic dataset labeling and noise conditions (a trace is positive if the condition is satisfied).}
\label{tab:noise_conditions}
\end{table*}

Concerning the real-life datasets, we used five different real-life event logs. 
By labeling each log with different labelings, we finally extracted $7$ different datasets. We describe in the following the real-life event logs and the related labelings used in the evaluation, which are summarised in \tablename~\ref{table:labellings}.

\textbf{\bpicEleven.} The BPI Challenge 2011~\cite{bpichallenge2011} pertains to a healthcare process related to the treatment of patients diagnosed with cancer in a large Dutch academic hospital. 
 Each trace refers to the treatment of a different patient. The event log contains domain specific attributes -- both trace attributes and event attributes. For instance, \act{Age}, \act{Diagnosis}, and \act{Treatment\ code} are trace attributes and \act{Activity\ code}, \act{Number\ of\ executions}, \act{Specialism\ code}, \act{Producer\ code}, and \act{Group} are event attributes. We use two different temporal properties for labeling the event log: the first one ($\phi_{11}$) checks if eventually the activity \act{tumor\ marker\ CA-19-9}, or the activity \act{ca-125\ using\ meia} occur; the second labeling ($\phi_{12}$) checks whether the activity \act{CEA - tumor\ marker\ using\ meia} is eventually followed by the activity \act{squamous\ cell\ carcinoma\ using\ eia}.

\textbf{\bpicFifteen.} The BPI Challenge 2015~\cite{bpichallenge2015} collects the event logs from five Dutch municipalities, pertaining to a building permit application process. The event logs, for each of the five municipalities, are denoted as \bpicFifteen\_i, where i = 1 $\ldots$ 5 indicates the municipality number. We consider, for our experiments, the event logs from two municipalities (\bpicFifteen\_3 and \bpicFifteen\_4). We apply to both the same labeling based on a temporal property ($\phi_{21}$) indicating whether activity \act{send\ confirmation\ receipt} is eventually followed by activity \act{retrieve\ missing\ data}.

\textbf{\production.} This event log~\cite{production} contains data from a manufacturing process. Each trace records information about activities, workers and/or machines involved in producing an item. The labeling ($\phi_{31}$) is based on whether or not the number of rejected work orders is higher than zero.

\textbf{\sepsis.} The Sepsis event log~\cite{sepsis} records trajectories of patients with symptoms of the sepsis condition in a Dutch hospital. This log contains execution traces related to the treatment of sepsis cases. Each case logs events from the patient's registration in the emergency room to the discharge of the patient from the hospital. Among others, laboratory tests together with their results are recorded as events. 
 Two different labelings are considered for this event log: the first one ($\phi_{41}$) checks whether the patient returns to the emergency room within $28$ days from the discharge; the second labeling ($\phi_{42}$) checks whether the patient is discharged from the hospital on the basis of a reason different from \act{Release\ A}, which is the most common release type.

\setlength{\tabcolsep}{6pt}

\begin{table}[t]
	\centering
\scalebox{.6}{
\begin{tabular}{l c l}
\toprule
\textbf{Log} & \textbf{Labeling} & \textbf{Condition} \\
\midrule
\multirow{2}{*}{\bpicEleven} & $\phi_{11}$ & \decp{existence}(\act{tumor\ marker\ CA-19-9}) or \decp{existence}(\act{ca-125 using\ meia})  \\ \cmidrule{2-3}
& $\phi_{12}$ &  \decp{response}(\act{CEA - tumor\ marker\ using\ meia}, \act{squamous\ cell\ carcinoma\ using\ eia}) \\ \midrule

\bpicFifteen\_3 & \multirow{2}{*}{$\phi_{21}$} &  \multirow{2}{*}{\decp{response}(\act{send\ confirmation\ receipt},\act{retrieve\ missing\ data})} \\ 
\bpicFifteen\_4 & 	&  \\ \midrule

\production & $\phi_{31}$	&  number of rejected work orders $> 0$ \\ \midrule

\multirow{3}{*}{\sepsis} & \multirow{1}{*}{$\phi_{41}$} & The patient returns to the emergency room within 28 days from the discharge \\ \cmidrule{2-3}

& \multirow{2}{*}{$\phi_{42}$}	&  The patient is discharged from the hospital on the basis of something \\
                  &                 & other than \act{Release\ A}\\
\bottomrule
\end{tabular}
}
\caption{Dataset labelings (a trace is positive if the condition is satisfied).}
\label{table:labellings}
\end{table}

Finally, \tablename~\ref{table:datasets} reports the characteristics of the datasets (both synthetic and real-life), the specific prefix lengths used in the evaluation (selected as explained in the next section), as well as the distributions of positive and negative labels.

\begin{table}[t]
	\centering
\scalebox{.7}{
\begin{tabular}{l c l l c l l}
\toprule
\multirow{2}{*}{\textbf{Log}} & \multirow{2}{*}{\textbf{Labeling}} & \multirow{2}{*}{\textbf{Trace \#}} & \multirow{2}{*}{\textbf{Activity \#}} & \textbf{Prefix} &\textbf{Positive} & \textbf{Negative} \\
& & &  & \textbf{length} & \textbf{trace \%} & \textbf{trace \%} \\
\midrule
\multirow{3}{*}{\claim} &  $\phi_{S1}$ & \multirow{3}{*}{$4800$} &  \multirow{3}{*}{$17$} & 4 & $29.8\%$ & $70.2\%$ \\
                                      &  $\phi_{S2}$ &                       &                      &        7           & $26.2\%$ & $73.8\%$ \\
                                      &  $\phi_{S3}$ &                       &                      &        7           & $29.4\%$ & $70.6\%$ \\
\midrule
\multirow{3}{*}{\bpicEleven} & $\phi_{11}$ & \multirow{3}{*}{$1\,140$} & \multirow{3}{*}{$623$} & \multirow{3}{*}{6} & $40.2\%$ & $59.8\%$ \\ \cmidrule{6-7}
& $\phi_{12}$ &  &  &  & 78.3\% & 21.7\% \\ \midrule
\bpicFifteen\_3 & \multirow{3}{*}{$\phi_{21}$} & $1\,328$ & 380 & $36$ & 19.7\% & 80.3\% \\ \cmidrule{3-7}
\bpicFifteen\_4 &  & $577$ & $319$ & $36$ & 15.9\% & 84.1\% \\ \midrule
\production & $\phi_{31}$	& $220$ & $26$ & $5$ & 53.2\% & 46.8\% \\ \midrule
\multirow{3}{*}{\sepsis} & $\phi_{41}$ & \multirow{3}{*}{782} & \multirow{3}{*}{15} & \multirow{3}{*}{10} & 14.2\% & 85.8\%\\ \cmidrule{6-7}
&  $\phi_{42}$ &  &  &  & 85.8\% & 14.2\% \\
\bottomrule
\end{tabular}
}
\caption{Dataset statistics}
\label{table:datasets}
\end{table}

\subsection{Experimental setting}
\label{ssec:setting}
In order to evaluate the proposed approach, we applied the following procedure to each dataset:
\begin{itemize}
	\item we split the dataset in a training set ($48\%$), a validation set ($16\%$), a feedback set ($16\%$) and a testing set ($20\%$);
	\item we used the training set for building a Random Forest classifier and the validation set to optimize its hyperparameters \cite{DBLP:conf/caise/Francescomarino16} - in particular, the hyperparameter procedure was set to run $50$ trials optimizing the macro-F1 metrics;
	\item we used the feedback set for extracting the confusion matrix;
	\item we applied the proposed pipeline to select the relevant pairs of \FEIs;
  \item we shuffled the training set according to the selected pairs of \FEIs and retrained the classifier;
	\item we used the testing set for computing the accuracy of the classifier trained with the shuffled training set (and compared the results with the ones obtained with the original classifier, i.e., the one trained with the original training set).
\end{itemize}

For the synthetic datasets, we set the threshold related to the considered SHAP explanations $t$ to $10$ and the filter threshold $k$ to 3 - considering the fact that the data is not very sparse. In addition, we selected the prefix lengths used for the experiments (Table~\ref{table:datasets}) in order to minimize the amount of information used to train the classifier and to avoid data leaking. For the real-life datasets, we set the threshold related to the considered SHAP explanations $t$ to $10$ and we considered three values for the filter threshold $k$: $3$, $5$ and $10$. In addition, we selected, for each dataset, as prefix length (Table~\ref{table:datasets}) the first quintile of the trace lengths of the corresponding log (training+validation).
As trace encodings, we used the \simple encoding, the \complex encoding and the \declare encoding (described in Section~\ref{sec:declareEncoding}).  
For all the datasets, we set the \textit{support} threshold of the \declare encoding to $0.25$, in order to have a sufficient number of temporal patters that could precisely describe the trace behaviors. 

In order to evaluate the accuracy of the predictions, given the imbalance of some of the considered datasets (see Table~\ref{table:datasets}), we compute the macro-averaged F1, that is the average F1-score of the two classes.

\section{Results}
\label{sec:results}
In this section, we first report the results related to the synthetic datasets, then the ones related to the real-life datasets 
and, finally, we conclude with a discussion and some threats to the validity of the evaluation.
\subsection{Synthetic datasets}
	For the synthetic datasets, we first focus on the qualitative characterization of the wrong predictions and compare it to the expected characterization. We then report the results obtained on the testing sets when re-training the predictive model.

		\begin{table}[h!]
		\centering
	\scalebox{.7}{
	\begin{tabular}{l l l l}
	\toprule
	\textbf{Label.} & \textbf{Encoding} & \textbf{Set} & \textbf{Mined \FEIs}\\ 
	\midrule
	\multirow{8}{*}{$\phi_{S1}$} & \multirow{8}{*}{\simple} & \multirow{8}{*}{$F_N$} & 
	
	    $\pi_{\act{act}}(e_4)$ = \act{Accept Claim} \\

	\cmidrule{4-4}
		& & & $\pi_{\act{act}}(e_4)$ = \act{Accept Claim} and\\

	& & & $\pi_{\act{act}}(e_3)$ = \act{Low Insurance Check} \\
	
	\cmidrule{4-4}
	
	& & &  $\pi_{\act{act}}(e_4)$= \act{Accept Claim} and\\

	& & & $\pi_{\act{act}}(e_3)$ = \act{Low Insurance Check} and\\
	& & & $\pi_{\act{act}}(e_2)$ = \act{Low Medical History} \\

	\midrule
	
	\multirow{3}{*}{$\phi_{S2}$}  &  \multirow{3}{*}{\complex} & \multirow{3}{*}{$F_N$}  & 
									   \att{CType}=\val{Silver} \\

	\cmidrule{4-4}
								   &                             & 	&
								   \att{CType}=\val{Silver} and $\pi_{\act{act}}(e_2)$ = \act{Create Questionnaire} \\ 
	\cmidrule{4-4}
								   &                             &  &
	\att{CType}=\val{Silver} and \att{PClaims}=\val{No} \\ 
	\midrule
	
\multirow{11}{*}{$\phi_{S3}$} & \multirow{11}{*}{\declare} & \multirow{11}{*}{$F_N$} &
	    !(\decp{response(\act{Create Questionnaire}, \act{Accept Claim})}) and\\
	& & & !(\decp{response(\act{Create Questionnaire}, \act{Low Medical History})}) and\\

	& & & !(\decp{precedence(\act{Create Questionnaire}, \act{Accept Claim})}) \\

	\cmidrule{4-4}

	& & & !(\decp{response(\act{Create Questionnaire}, \act{Accept Claim})}) and\\
	& & & !(\decp{response(\act{Create Questionnaire}, \act{Low Medical History})}) and\\

	& & & !(\decp{precedence(\act{Create Questionnaire}, \act{Accept Claim})}) and\\
	& & & !(\decp{precedence(\act{Create Questionnaire}, \act{Low Medical History})}) \\

	\cmidrule{4-4}

	& & & !(\decp{response(\act{Create Questionnaire}, \act{Accept Claim})}) and \\
	& & & !(\decp{response(\act{Create Questionnaire}, \act{Low Medical History})}) and\\

	& & & !(\decp{precedence(\act{Create Questionnaire}, \act{Low Medical History})}) \\
	
	\bottomrule
	\end{tabular}
	}
	\caption{Explanation: synthetic results.}
	\label{tab:synthetic_explanation}
	\end{table}

Table~\ref{tab:synthetic_explanation} reports the mined \FEIs characterizing the wrong predictions.\footnote{By construction, for the considered labelings, only false negatives exist. We report the qualitative evaluation only of the top three characterizations.} In the case of $\phi_{S1}$, according to the conditions in Table~\ref{tab:noise_conditions}, we would expect that the incorrect predictions are the ones for which \act{AcceptClaim} occurs but not as the fifth event. Indeed, the model learns that only traces with \act{AcceptClaim} at position 5 are positive, while, according to the test set, all traces in which \act{AcceptClaim} occurs are positive, thus resulting in false negative predictions (whenever \act{AcceptClaim} occurs but not at position 5). All the three characterizations obtained for the $F_N$ set, indeed, contain the occurrence of the activity \act{AcceptClaim} at position 4, which is possibly the most frequent position in the log in which \act{AcceptClaim} occurs, apart from position 5. This condition is accompanied, in the characterization of this set of false negatives, by the activities \act{Low Medical History} and \act{Low Insurance Check} which, by design, strongly correlate with \act{Accept Claim}.
	
In the case of $\phi_{S2}$, according to the conditions in Table~\ref{tab:noise_conditions}, we would expect that the incorrect predictions are the ones for which \att{Age} is lower than 60 and \att{CType} is \val{Silver} (false positives). By looking at the three obtained characterizations of the false positive set,  all contain the condition \att{CType}=\val{Silver}, accompanied by \att{PClaims}=\val{No} and by the occurrence of the activity \act{CreateQuestionnaire} in the second position, conditions that often happen together with \att{CType}=\val{Silver}.
	
In the case of $\phi_{S3}$, according to the conditions in Table~\ref{tab:noise_conditions}, we expect that the false negatives are those traces for which \act{AcceptClaim} occurs and \decp{!(response(\act{Low Medical History}, \act{Create Questionnaire}))} is not satisfied, which means that either \act{LowMedicalHistory} does not occur, or it occurs and it is followed by \act{CreateQuestionnaire}. The obtained results, indeed, suggest that the characterization of the set of the false negatives is given by \decp{!(response(\act{Create Questionnaire}, \act{Low Medical History}))}, that means that \act{Create Questionnaire} occurs (and this is true by design in every trace of the log), and \act{Low Medical History} cannot occur after (that is either it does not occur or it occurs and it is followed by \act{Create Questionnaire}). The other conditions discovered together with this common behavior either make this behavior stronger (\decp{!(precedence(\act{Create Questionnaire}, \act{Low Medical History}))} means that that \act{Low Medical History} occurs always and it is followed by \act{Create Questionnaire}), or ensure that \act{Accept Claim} occurs (\decp{!(precedence(\act{Create Questionnaire}, \act{Accept Claim}))} means that \act{Accept Claim} occurs and it is followed by \act{Create Questionnaire}).

Table~\ref{tab:synthetic_results} (column \emph{Baseline}) reports, for the synthetic datasets, the values of the macro-averaged F1 metrics obtained by evaluating the models trained with the original training set on the testing set with the three encodings. The column \emph{re-training}, in the same table, reports the values of the macro-averaged F1 metrics related to the evaluation of the predictive model retrained with the shuffled data. 
In particular, for each encoding and corresponding labeling, we selected the most important \FEIs and we retrained the classifier with a dataset in which we reduce the occurrence of such patterns of features. By looking at the table, in which the best result per dataset (and labeling) is reported in bold, we can observe that the re-training procedure always has a positive effect on the results, independently of the encoding used. However, the largest improvement is obtained with the \complex encoding followed by the \simple and finally by the \declare encoding.
	\begin{table}[h] 
		\centering
		\scalebox{.8}{
	\begin{tabular}{c l c c}
	\toprule
	\textbf{Labeling} & \textbf{Encoding} & \textbf{Baseline} & \textbf{re-training} \\
	\midrule
	 $\phi_{S1}$ & \simple  & 0.58 & \textbf{0.75} \\ 
	 $\phi_{S2}$ & \complex & 0.76 & \textbf{0.96} \\ 
	 $\phi_{S3}$ & \declare & 0.59 & \textbf{0.70} \\
	\bottomrule
	\end{tabular}
	}
	\caption{Results related to the synthetic datasets.}
	\label{tab:synthetic_results}
	\end{table}

\subsection{Real-life datasets}
Table~\ref{table:top3and5and10} (column \emph{Baseline}) reports, for the real-life datasets, the values of the macro-averaged F1 metrics obtained by evaluating the models trained with the original training set on the testing set with different encodings. 

The table, in which the results related to the best encoding(s) per dataset are reported in italic, shows that none of the encodings is the best one over all the datasets. For the two labelings of \sepsis and for \bpicEleven $\phi_{21}$, no significant differences can be observed among the encodings (the difference is always $\leq 0.04$). Concerning the remaining cases, for \bpicEleven $\phi_{22}$, the \complex encoding seems to be the best one, for the two \bpicFifteen\xspace datasets, the \simple and the \complex encodings outperform the \declare encoding, while, for the \production dataset, the \declare encoding is the best one. By looking at the characteristics of the datasets, it seems that the \declare encoding performs better on balanced datasets (e.g., the \production dataset), 
while the \simple and \complex encodings seem to be more suitable to imbalanced datasets with a large number of activities (e.g., the \bpicFifteen\xspace datasets). 

These observations highlight that there is not a unique encoding that best suits all datasets (\textbf{RQ1}).

\begin{table}[h] 
	\centering
\scalebox{.7}{
\begin{tabular}{l c l c@{\hskip 0.2in} c@{\hskip 0.2in} c@{\hskip 0.2in} c}
\toprule
\multirow{2}{*}{\textbf{Dataset}} & \multirow{2}{*}{\textbf{Labelling}} & \multirow{2}{*}{\textbf{Encoding}} & \multirow{2}{*}{\textbf{Baseline}} & \multicolumn{3}{c}{\textbf{re-training}} \\
 &  &  &  & $\mathbf{k=3}$ & $\mathbf{k=5}$ & $\mathbf{k=10}$ \\
\midrule
\multirow{6}{*}{\bpicEleven} &  \multirow{3}{*}{$\phi_{11}$}
 & \simple  & \textit{0.57}  &                 0.57   &                 0.58   &         \textbf{0.59}  \\
 & & \complex &         0.56  & \textbf{\textit{0.62}} & \textbf{\textit{0.62}} &         \textit{0.61}  \\  
 & & \declare   & \textit{0.57} &                 0.59   &                 0.58   &         \textbf{0.60}  \\ \cmidrule{2-7}
 & \multirow{3}{*}{$\phi_{12}$} 
 & \simple  &         0.47  &                 0.54   &                 0.46   &         \textbf{0.58}  \\ 
 & & \complex & \textit{0.55} & \textit{0.77}          & \textbf{\textit{0.82}} &         \textit{0.75}  \\ 
 &  & \declare   &         0.41  &                 0.47   &         \textbf{0.48}  &                 0.46   \\\midrule
\multirow{3}{*}{\bpicFifteen\_3} & \multirow{6}{*}{$\phi_{21}$}
 & \simple  &         0.89  &                 0.92   & \textbf{\textit{0.94}} &                 0.91   \\
 & & \complex & \textit{0.92} &         \textbf{0.94}  &                 0.93   &                 0.89   \\ 
 & & \declare   &         0.43  & \textbf{\textit{0.97}} &                 0.43   & \textbf{\textit{0.97}} \\ \cmidrule{3-7}
\multirow{3}{*}{\bpicFifteen\_4} & 
 & \simple  & \textit{0.83} &         \textit{0.90}  &         \textit{0.89}  & \textbf{\textit{0.92}} \\
 & & \complex &         0.81  &         \textbf{0.85}  &         \textbf{0.85}  &                 0.84   \\ 
 & & \declare   &         0.74  &                 0.83   &         \textbf{0.84}  &         \textbf{0.84 } \\ \midrule
\multirow{3}{*}{\production} & \multirow{3}{*}{$\phi_{31}$}
 & \simple  &         0.36  &         \textbf{0.41}  &         \textbf{0.41}  &         \textbf{0.41}  \\
 & & \complex &         0.40  &                 0.42   &         \textbf{0.47}  &                 0.44   \\ 
 & & \declare   & \textit{0.52} &\textbf{ \textit{0.60}} &\textbf{ \textit{0.60}} & \textbf{\textit{0.60}} \\ \midrule
\multirow{6}{*}{\sepsis} & \multirow{3}{*}{$\phi_{41}$}
 & \simple  &         0.49  &                 0.50   & \textbf{\textit{0.52}} & \textbf{\textit{0.52}} \\
 & & \complex &         0.46  &                 0.46   &         \textbf{0.49}  &                 0.46   \\ 
 & & \declare   & \textit{0.50} & \textbf{\textit{0.56}} &                 0.46   &                 0.49   \\ \cmidrule{2-7}
 & \multirow{3}{*}{$\phi_{42}$}
 & \simple  & \textit{0.50} &         \textit{0.59}  &         \textbf{0.60}  &                 0.52  \\
 & & \complex &         0.46  &         \textbf{0.55}  &         \textbf{0.55}  &         \textit{0.54} \\
 & & \declare   &         0.48  &                 0.51   & \textbf{\textit{0.64}} &                 0.51  \\
\bottomrule
\end{tabular}
}
\caption{F1 measure values related to training (column \emph{Baseline}) and re-training (column \emph{re-training}) for $k=3$, $k=5$ and $k=10$.}
\label{table:top3and5and10}
\end{table}

Table~\ref{table:top3and5and10} also reports the results obtained by applying the shuffling procedure described in Section~\ref{sec:approach} to the training set with $k=3$, $k=5$ and $k=10$ (column \emph{re-training}). Also in this case, the results related to the best encoding per dataset are highlighted in italic, while the best result(s) per encoding are reported in bold.

By focusing on the column related to $k=3$ and looking at the difference in terms of performance between training and re-training, 
we can observe that the re-training procedure is able to achieve either the same results obtained with the original training, or an improvement on all the datasets. Only in two cases, i.e., the \simple encoding for \bpicEleven $\phi_{11}$ and the \complex encoding for \sepsis $\phi_{41}$, the results obtained with the re-training procedure are exactly the same as the ones obtained with the original training dataset. In all other cases, there is an improvement of the F1-measure that ranges from very low values ($\sim 0.01$), up to $0.54$ for the \declare encoding of \bpicFifteen\_3.
By looking at the improvement obtained per type of encoding, we can notice that, overall, the \declare encoding is able to obtain better improvements than the \complex and the \simple encodings: in four out of seven cases, the \declare encoding obtains the best improvement. 
The overall best improvements are obtained with the \declare encoding for \bpicFifteen\_3 and the \complex encoding for \bpicEleven $\phi_{12}$ ($+0.54$ and $+0.22$, respectively).

Also in the case of $k=5$, the re-training procedure improves or does not worsen the results obtained with the original training, except for \bpicEleven $\phi_{12}$ (\simple encoding) and \sepsis $\phi_{41}$ (\declare encoding), where we observe a slight decrease of the F1-measure, and \bpicFifteen\_3 (\declare encoding), where the F1-measure does not change.
The results obtained with the re-training procedure with $k=5$ are very close to the ones related to $k=3$, except for \bpicEleven $\phi_{12}$ showing a decrease in the performance for the \simple encoding, \bpicFifteen\_3 with an important decay in the performance for the \declare encoding, and \sepsis $\phi_{41}$ and $\phi_{42}$, showing, respectively, a deterioration and an improvement of the performance for the \declare encoding. Concerning the comparison between the encodings, differently from the results obtained with $k=3$, the results obtained with $k=5$ show that the \complex encoding is able to obtain better improvements with respect to the other encodings.

Finally, the results related to $k=10$ also reveal that, overall, the re-training procedure improves or preserves the results obtained with the original training - except for the \complex encoding for \bpicFifteen\_3 and for the \declare encoding for \sepsis $\phi_{41}$, for which the results obtained with the re-training procedure are worse than the ones obtained with the original training. Also in this case, the obtained results are very close to the ones obtained with $k=3$, except for the \declare encoding for \sepsis $\phi_{41}$, whose results are closer to the ones obtained with $k=5$, and the \simple encoding for \sepsis $\phi_{42}$.
In this case, no main differences can be identified in terms of best encoding per dataset - except for \bpicEleven $\phi_{12}$ (for which the \complex encoding is the best one), and \bpicFifteen\_3 and \production (for which the \declare encoding is the best one). 

By considering the results obtained with both the synthetic and the real-life datasets (for $k=3$), we can state that, overall, the re-training procedure is able to enhance or, in the worst case, preserve the results obtained with the original training (\textbf{RQ2}). The results obtained obtained with the real-life datasets for $k=5$ and $k=10$ are not too far from the ones obtained with $k=3$ but, for this values, sometimes the re-training procedure affects the F1-measure negatively. We do not report the results obtained with $k>10$, as, for those values, the improvements in the results obtained after the re-training procedure start decreasing.

\subsection{Threats to validity}
The main threat to the validity of the evaluation carried out is related to its generalizability. Indeed, we limited the evaluation of the proposed approach to 7 datasets. However, this threat is mitigated by the fact that the datasets are real-life datasets belonging to different domains, with different characteristics (number of traces in the log, number of events in the traces, number of activities) and labeling distributions. A second threat is related to the internal validity. Indeed, we fixed the SHAP threshold to $t=10$ and we only reported the results related to $k=3$, $k=5$ and $k=10$. However, in both cases, we selected the values based on some empirical evidence. Specifically, for $k>10$, we observed a decay in the performance of the re-training procedure.

\section{Related Work}
\label{sec:related}
To the best of our knowledge, no other works exist on the accuracy improvement exploiting frequent explanations, except for our previous work \cite{DBLP:conf/bpm/RizziFM20}. We hence first position our work with respect to the works in \ppm,
the works focused on explanations in \ppm, 
and the works improving the predictive model's performance employing (user) feedback; we finally address the specific comparison with our previous work~\cite{DBLP:conf/bpm/RizziFM20}.

Among the different prediction tasks in the \ppm state-of-the-art, an important group of papers focuses, as this work, on predicting outcomes~\cite{DBLP:journals/tkdd/TeinemaaDRM19}. For instance, several approaches deal with predicting the fulfilment (or the violation) of a boolean predicate in a running case~\cite{DBLP:conf/caise/MaggiFDG14,DBLP:journals/tkdd/TeinemaaDRM19}. Initial approaches and encodings for outcome-oriented \ppm have been enhanced in terms of performance by introducing pre-processing steps based on clustering and bucketing~\cite{DiFrancescomarinoetal2017,DeLeoni2016}, as well as in terms of prediction accuracy by introducing different types of encodings~\cite{DBLP:conf/bpm/LeontjevaCFDM15,DBLP:conf/bpm/LakshmananDKCK10}. Moreover, more recently, deep learning approaches have also been investigated for predicting outcomes~\cite{DBLP:journals/bise/KratschMRS21,PARK2020113191}.
Differently from these works, the focus of this paper is not on proposing a specific outcome prediction method, but rather on explaining why predictive models are wrong and on leveraging these explanations for eventually improving the performance of the predictive model.

A number of works have recently focused on providing explanations in \ppm~\cite{DBLP:conf/ecis/StierleBWZM021}. Some of them focus on interpretable predictive models~\cite{DBLP:conf/bpm/PauwelsC20,DBLP:conf/caise/BohmerR20}, others on model-specific approaches~\cite{harl2020explainable,DBLP:conf/bpm/SindhgattaMOB20,DBLP:journals/ki/RehseMF19,DBLP:journals/corr/abs-2008-07993,DBLP:conf/icpm/Pasquadibisceglie21}, while a third group provides explanations for model agnostic predictive models~\cite{DBLP:conf/icsoc/SindhgattaOM20,Mehdiyev2021,galanti_explainable_2020,DBLP:conf/icpm/HsiehMO21,DBLP:conf/emcis/HuangMP21}. The approach presented in this paper belongs to this last class of works.
Many of the works in this group leverage standard post-hoc explainers such as LIME~\cite{DBLP:journals/corr/RibeiroSG16} and SHAP~\cite{DBLP:conf/nips/LundbergL17}. For example, in~\cite{DBLP:conf/icsoc/SindhgattaOM20}, LIME is applied to XGBoost classifiers, while, in~\cite{galanti_explainable_2020}, SHAP values are leveraged for providing users with explanations in the form of tables explaining the predictions related to a specific ongoing process execution. In~\cite{Mehdiyev2021}, instead, a novel approach for explaining the predictions returned by a deep learning classifier is proposed. Counterfactual explanations, obtained by adapting counterfactual explainers such as DiCE~\cite{DBLP:conf/fat/MothilalST20} and LORE~\cite{DBLP:journals/expert/GuidottiMGPRT19} are instead investigated in~\cite{DBLP:conf/icpm/HsiehMO21} and \cite{DBLP:conf/emcis/HuangMP21}, respectively.
In this paper, we focus on the usage more than on the computation of explanations: we indeed leverage prediction explanations (returned by SHAP) in order to understand why predictors are wrong and improve their accuracy.
Some works in machine learning focus on the enhancement of the accuracy of a predictive model by leveraging users' feedback. For instance, in~\cite{DBLP:conf/iui/StumpfRLBDSDH07}, the authors show
that machine learning systems can explain their reasoning to users and that users, in turn, can provide informative feedback to the learning system, thus effectively improving it. The works in~\cite{DBLP:conf/iui/ChklovskiRG05,DBLP:conf/iui/McCarthyRMS05} show that semi-formal types of feedback are preferred by users to feedback based on editing feature-value pairs. 
A key difference between these approaches and the approach presented in this paper is that our approach aims at automatically discovering the feedback and applying it to the predictive models without any user intervention.

In~\cite{DBLP:conf/bpm/RizziFM20}, an approach for explaining why a predictive model for outcome-oriented prediction provides wrong predictions and for eventually improving its accuracy is presented. The approach is semi-automatic, leaving to the user the burden of selecting the most suitable explanations to leverage. Moreover, the work only focuses on state-of-the-art \ppm encodings~\cite{DBLP:conf/bpm/LeontjevaCFDM15}. Instead, in this work, we provide a completely automated procedure that does not require any human intervention for the re-training of the predictive models, we compare state-of-the-art \ppm encodings with a declarative encoding (the \declare encoding) and we extend the evaluation to different real-life datasets.

\section{Conclusions}
\label{sec:conclusions}
In this paper, we (i) employ a novel \ppm encoding for explaining and characterizing (wrong) predictions in terms of temporal patterns; and (ii) introduce a completely automated pipeline for the identification of the most discriminative features inducing a predictor to make mistakes, the shuffling of the training set and the re-training of the predictor to improve its accuracy. Moreover, we compare the declarative encoding with state-of-the-art \ppm encodings both in terms of accuracy and in terms of effectiveness of the re-training procedure using synthetic and real-life datasets. The results show that the re-training strategy is able to return - in a completely automated way - better improved or no-worse results than the original training. Moreover, although no major differences exist among the encodings in terms of accuracy of trained and re-trained models, the \declare encoding performs better on balanced datasets, while the \simple and \complex encodings seem to be more suitable to imbalanced datasets with a large number of activities. 

As future work, we plan to investigate the usage of encodings that combine the \declare and the \complex encoding. In particular, we would like to consider features based on ``data-aware'' temporal declarative patterns. Moreover, we would like to investigate how to equip the proposed pipeline with techniques for the optimization of the hyperparameters $t$ and $k$. Finally, we would like to apply the proposed pipeline to multi-class prediction problems by defining and leveraging different strategies for dealing with a multi-class confusion matrix, ranging from the simple clustering of classes in only two clusters (the ``good'' and the ``bad'' one) to more complex and fine-grained mechanisms for dealing with the poorly populated quadrants of a multi-class confusion matrix.

\ifCLASSOPTIONcaptionsoff
  \newpage
\fi



\bibliographystyle{IEEEtran}
%

\bibliography{bibliography}

%

    \vspace{-13mm}
    \begin{IEEEbiography}[{\includegraphics[width=1in,height=1.25in,clip,keepaspectratio]{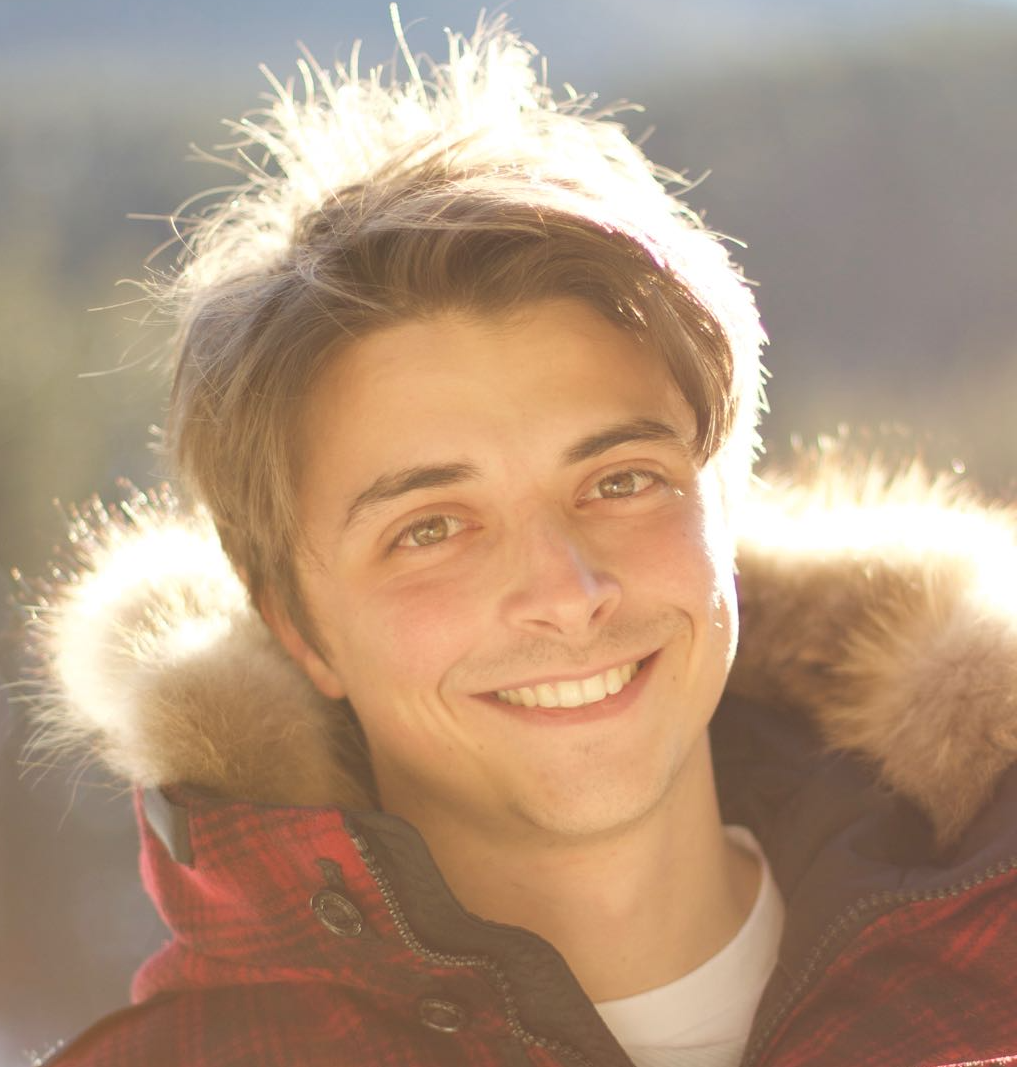}}]{Williams Rizzi}
    is currently a PhD student in Computer Science enrolled in a joint PhD Programme between Fondazione Bruno Kessler and Free University of Bozen-Bolzano, Italy. His research interests focus on the application of Machine Learning techniques to the Predictive Process Monitoring domain and he is currently actively developing Nirdizati, one of the state-of-the-art Predictive Process Monitoring tools. 
    \end{IEEEbiography}

    \vspace{-13mm}
    \begin{IEEEbiography}[{\includegraphics[width=1in,height=1.25in,clip,keepaspectratio]{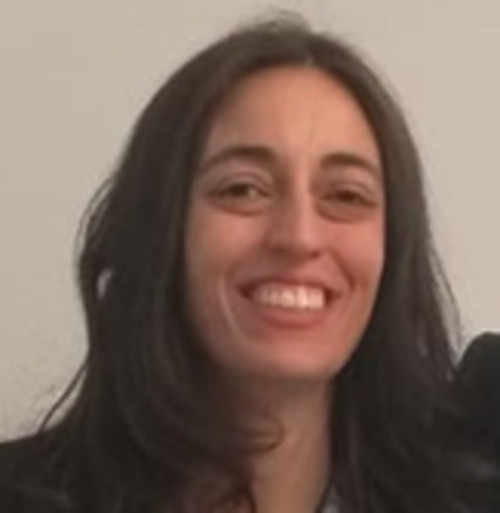}}]{Chiara Di Francescomarino}
      is a researcher at Fondazione Bruno Kessler (FBK) in the Process and Data Intelligence (PDI) Unit. She received her PhD in Information and Communication Technologies from the University of Trento, working on business process modeling and reverse engineering from execution logs. She is currently working in the field of process mining, investigating problems related to process monitoring, process discovery, as well as predictive process monitoring based on historical execution traces. 
      She serves as PC member in top conferences in the business process management field and as peer reviewer in international journals.
    \end{IEEEbiography}
    \vspace{-13mm}
    \begin{IEEEbiography}[{\includegraphics[width=1in,height=1.25in,clip,keepaspectratio]{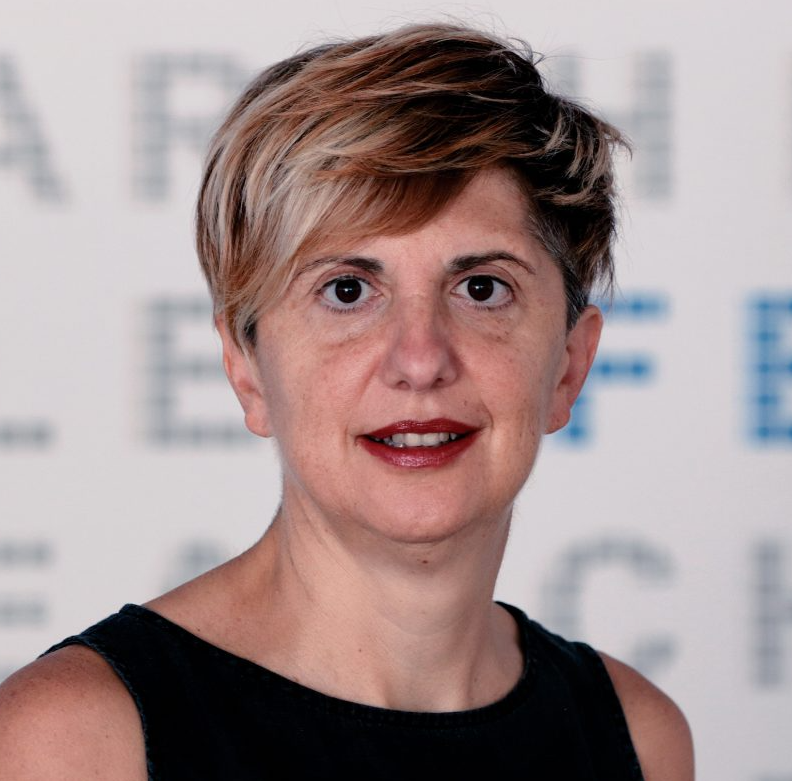}}]{Chiara Ghidini}
      is a senior Research Scientist at Fondazione Bruno Kessler (FBK), Trento, Italy, where she heads the Process \& Data Intelligence (PDI) research unit. Her scientific work in the areas of Semantic Web, Knowledge Engineering and Representation, Multi-Agent Systems and Process Mining is internationally well known and recognised, and she has published more than 100 papers in those areas. Dr. Ghidini has acted as PC or track chair in the organisation of workshops and conferences on multiagent systems, Contexts-based representations, Knowledge Engineering and Capturing, Semantic Web and Business Process Management. 
    \end{IEEEbiography}

    \vspace{-13mm}
    \begin{IEEEbiography}[{\includegraphics[width=1in,height=1.25in,clip,keepaspectratio]{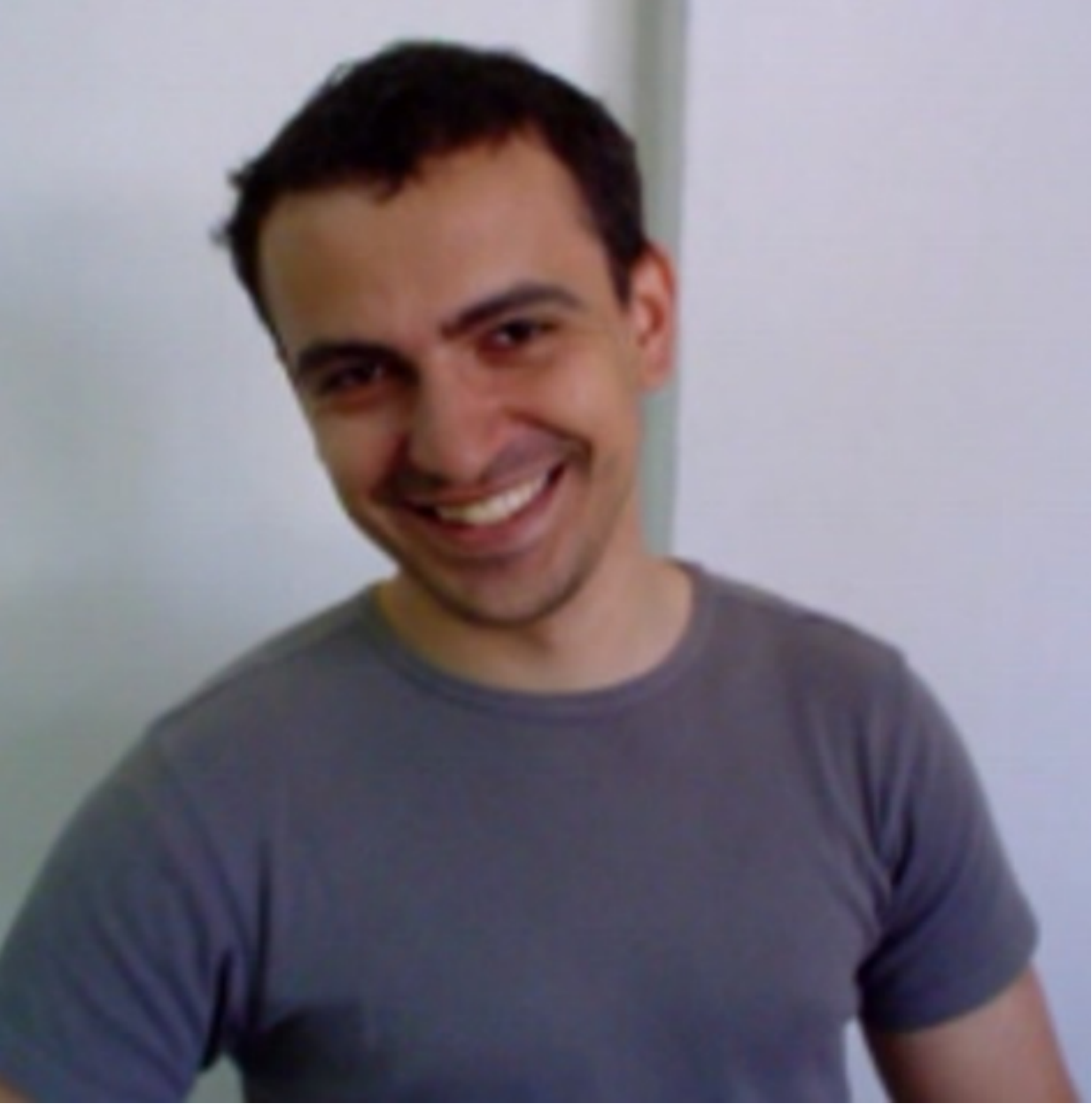}}]{Fabrizio Maria Maggi}
    is currently Associate Professor at the Research Centre for Knowledge and Data (KRDB) - Faculty of Computer Science - Free University of Bozen-Bolzano. His research interest has focused in the last years on the application of Artificial Intelligence to Business Process Management. He authored more than 150 articles on process mining, declarative process notations, predictive process monitoring. He serves as program committee member of the top conferences in the field of Business Process Management and Information Systems. 
    \end{IEEEbiography}




\end{document}